%% file: main.tex
\documentclass[10pt,twocolumn,letterpaper]{article}

\usepackage[pagenumbers]{cvpr} % To force page numbers, e.g. for an arXiv version
\usepackage{xcolor}
\usepackage{booktabs}
\definecolor{cvprblue}{rgb}{0.21,0.49,0.74}
\usepackage[pagebackref,breaklinks,colorlinks,allcolors=cvprblue]{hyperref}
\usepackage{algorithm}
\usepackage{algpseudocode}
\usepackage{mathabx}
\usepackage{makecell}
\usepackage{caption}
\usepackage{subcaption}
\usepackage[most]{tcolorbox}
\usepackage{textcomp}
\usepackage{cuted}

\newcounter{teasersubfig}                % sub-(a),(b) counter
\renewcommand{\theteasersubfig}{\alph{teasersubfig}}
\newcommand{\teasersubcap}[2]{% \teasersubcap{Caption text}{label}
  \refstepcounter{teasersubfig}%
  (\theteasersubfig)~#1\label{#2}%
}

\def\paperID{} % *** Enter the Paper ID here
\def\confName{CVPR}
\def\confYear{2026}

\title{Stable and Efficient Single-Rollout RL for Multimodal Reasoning}

\author{
Rui Liu$^{1,2^*}$,
Dian Yu$^{1}$,
Lei Ke$^{1}$,
Haolin Liu$^{3}$,
Yujun Zhou$^{4}$,
Zhenwen Liang$^{1}$, \\ 
Haitao Mi$^{1}$,
Pratap Tokekar$^{2}$,
Dong Yu$^{1}$ \\ 
$^{1}$Tencent AI Lab, Bellevue
$^{2}$University of Maryland, College Park \\
$^{3}$University of Virginia
$^{4}$University of Notre Dame \\
Project Page: \href{https://mssr-proj.github.io}{https://mssr-proj.github.io}
}

\begin{document}

\twocolumn[{%
\renewcommand\twocolumn[1][]{#1}%
\maketitle

\begin{center}
\captionsetup{type=figure}
\setcounter{teasersubfig}{0} % reset a,b for this teaser

\begin{minipage}{0.33\textwidth}
  \centering
  \includegraphics[height=4.4cm]{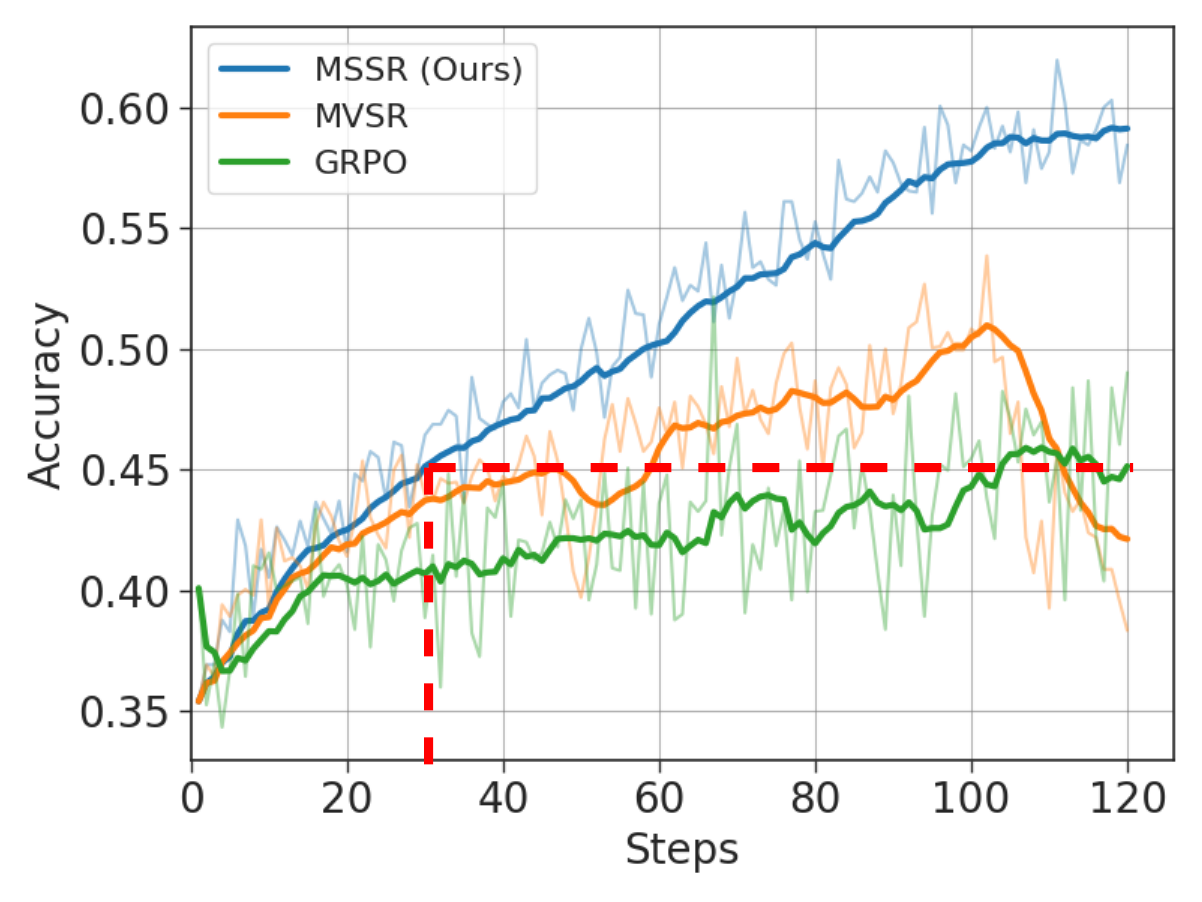}
  % \vspace{3pt}
  \teasersubcap{Training accuracy}{fig:train_acc}
\end{minipage}
\hfill
\begin{minipage}{0.33\textwidth}
  \centering
  \includegraphics[height=4.4cm]{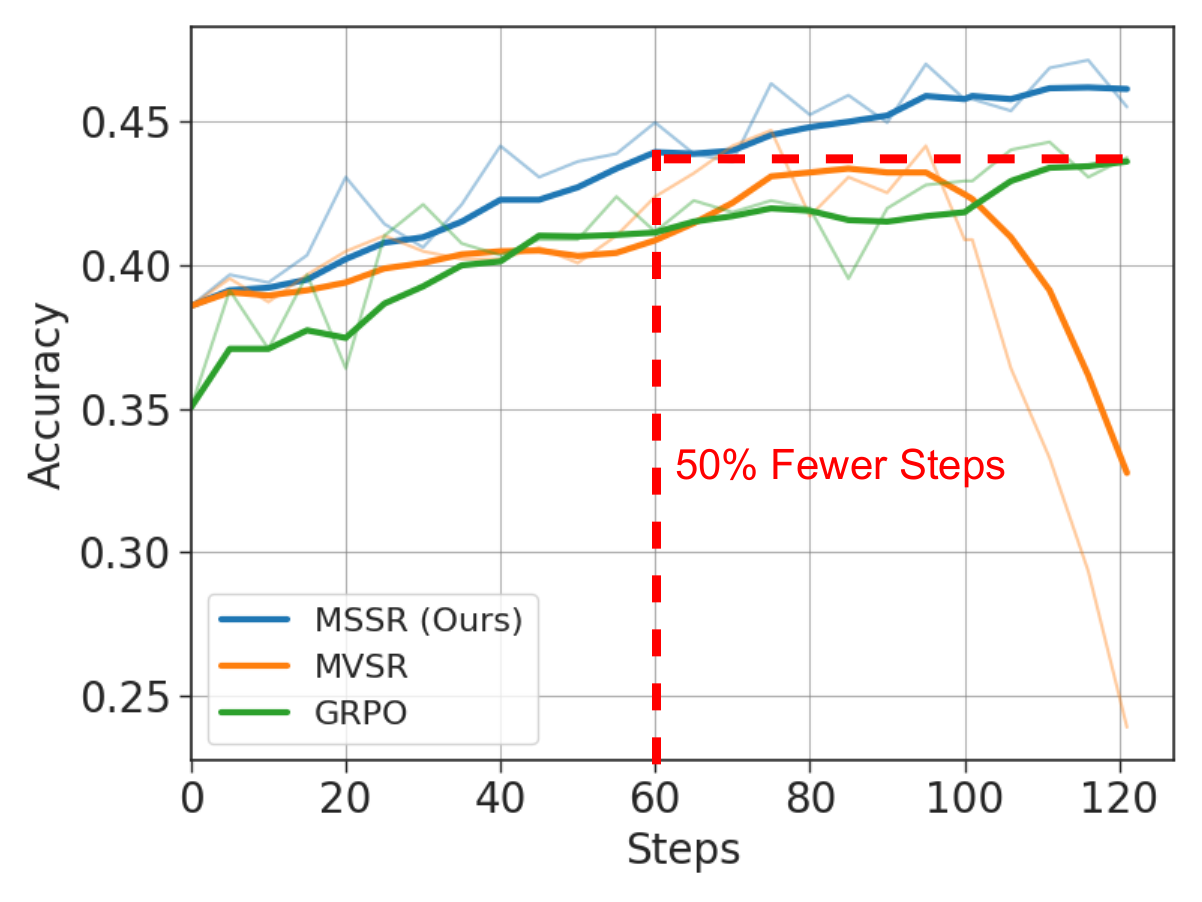}
  % \vspace{3pt}
  \teasersubcap{Validation accuracy}{fig:val_acc}
\end{minipage}
\hfill
\begin{minipage}{0.33\textwidth}
  \centering
  \includegraphics[height=4.4cm]{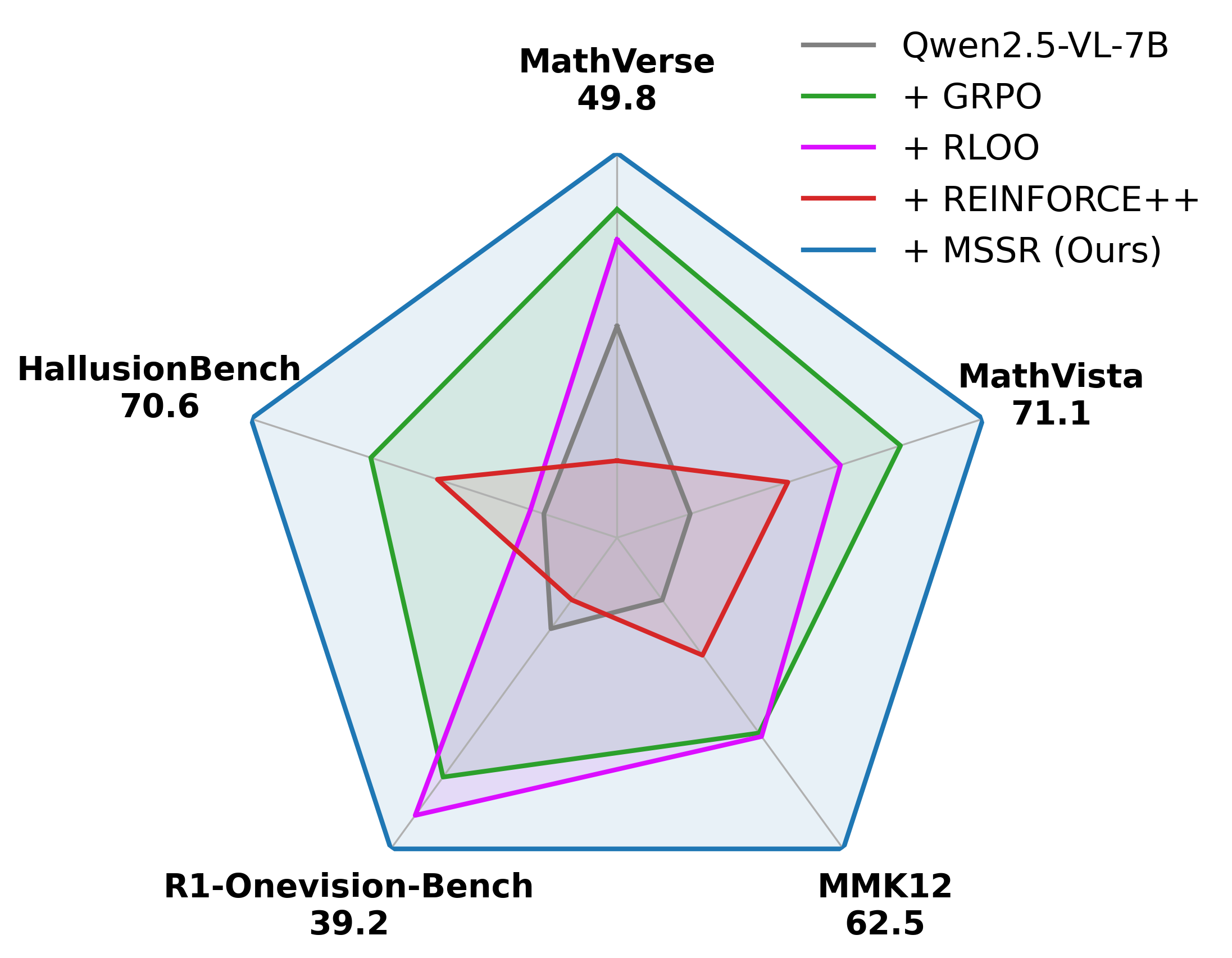}
  % \vspace{3pt}
  \teasersubcap{Generalization performance}{fig:radar}
\end{minipage}

\captionof{figure}{\textbf{Performance overview of MSSR:} 
 (a–b) Training and validation accuracy of MVSR (Multimodal Vanilla Single-Rollout), GRPO \cite{shao2024deepseekmath} and our MSSR, trained on the Vision-R1-RL \citep{huang2025vision} training set and validated on its corresponding validation set. MSSR remains stable and improves steadily, whereas MVSR is unstable and collapses. Notably, MSSR reaches a similar final validation accuracy to GRPO with half of the training steps, highlighting its superior training compute efficiency. 
(c) Our MSSR achieves higher generalization performance across diverse multimodal reasoning benchmarks, including MathVerse \citep{zhang2024mathverse}, MathVista \citep{lu2023mathvista}, MMK12 \cite{meng2025mm}, R1-Onevision-Bench \cite{yang2025r1}, and HallusionBench \citep{guan2024hallusionbench}, compared to other baselines including GRPO \cite{shao2024deepseekmath}, RLOO \cite{ahmadian2024back}, and REINFORCE++ \cite{hu2025reinforce++}. For fair comparisons, we have equivalent total number of rollouts per step for all methods.}

\label{fig:page1}
\end{center}%
}]

\let\thefootnote\relax\footnotetext{* Work done during an internship at Tencent AI Lab, Bellevue, WA.}

\input{sec/abstract}

% \input{sec/intro}
\input{sec/intro_updated}

% \input{sec/related}
\input{sec/related_updated}
\input{sec/prelim}

\input{sec/approach}

\input{sec/experiments}

\input{sec/conclusions}

{
    \small
    \bibliographystyle{ieeenat_fullname}
    \bibliography{main}
}

\input{sec/suppl}

\end{document}

%% file: sec/abstract.tex
\begin{abstract}

\vspace{-10pt}

Reinforcement Learning with Verifiable Rewards (RLVR) has become a key paradigm to improve the reasoning capabilities of Multimodal Large Language Models (MLLMs). However, prevalent group-based algorithms such as GRPO require multi-rollout sampling for each prompt. While more efficient single-rollout variants have recently been explored in text-only settings, we find that they suffer from severe instability in multimodal contexts, often leading to training collapse. To address this training efficiency-stability trade-off, we introduce \textbf{MSSR} (Multimodal Stabilized Single-Rollout), a group-free RLVR framework that achieves both stable optimization and effective multimodal reasoning performance. MSSR achieves this via an entropy-based advantage-shaping mechanism that adaptively regularizes advantage magnitudes, preventing collapse and maintaining training stability. While such mechanisms have been used in group-based RLVR, we show that in the multimodal single-rollout setting they are not merely beneficial but essential for stability. In in-distribution evaluations, MSSR demonstrates superior training compute efficiency, achieving similar validation accuracy to the group-based baseline with half the training steps. When trained for the same number of steps, MSSR's performance surpasses the group-based baseline and shows consistent generalization improvements across five diverse reasoning-intensive benchmarks. Together, these results demonstrate that MSSR enables stable, compute-efficient, and effective RLVR for complex multimodal reasoning tasks.

\end{abstract}

% are computationally expensive
% MSSR achieves at least comparable accuracy in half the training steps and surpasses leading performance under comparable budgets. Moreover, MSSR-trained policies exhibit stronger generalization, showing consistent improvements across six diverse reasoning-intensive benchmarks. 

%% file: sec/intro_updated.tex
\vspace{-10pt}
\section{Introduction} \label{sec:intro}

Reinforcement learning (RL) fine-tuning has become central to aligning Large Language Models (LLMs) and Multimodal Large Language Models (MLLMs) with human preferences. Early efforts, such as Reinforcement Learning from Human Feedback (RLHF) \cite{ouyang2022training, bai2022training}, optimized models toward human-preferred behaviors, improving alignment and performance. More recently, attention has shifted toward Reinforcement Learning with Verifiable Rewards (RLVR) \cite{luong2024reft, lambert2024t, guo2025deepseek, zheng2025parallel, dai2025cde, zheng2025learning, dai2025r1}, which replaces human feedback with automatically verifiable correctness signals. These binary rewards enable models to learn directly from objective supervision and have been successfully applied to multimodal reasoning \cite{liu2025vogue, li2025self, huang2025vision, zhang2025r1, wang2025vl, Qwen-VL, deng2025openvlthinker}.

Despite this progress, multimodal RLVR methods still face key challenges. They typically require generating a group of rollouts for each input. Prevalent methods such as GRPO \cite{shao2024deepseekmath} rely on multiple rollouts per input to infer relative advantages, resulting in substantial cost on repeated forward passes through the vision and language encoders, which becomes particularly expensive for large multimodal models. Moreover, when all rollouts in a group yield identical outcomes (e.g., all correct or all incorrect), the relative advantage collapses to zero, producing no learning signal and reducing rollout utilization efficiency \cite{yu2025dapo}. This raises a key question: \textbf{Can multimodal RLVR be made both compute efficient and stable, without sacrificing or even improving accuracy?}

We introduce \textbf{MSSR} (\textbf{M}ultimodal \textbf{S}tabilized \textbf{S}ingle-\textbf{R}ollout), a group-free approach that requires only one rollout per multimodal input, aiming to achieve both high training compute efficiency and stable optimization. Although single-rollout RLVR has recently been explored in text-only settings \cite{xu2025single, brantley2025accelerating}, extending it to multimodal reasoning is considerably more challenging. The inclusion of dense, high-dimensional visual inputs in images substantially increases input variance and complicates cross-modal credit assignment~\citep{liu2025vogue, feng2025rewardmap}. Our experiments confirm naive single-rollout strategies that succeed in text-only RLVR fail to transfer to multimdoal settings, frequently resulting in unstable learning and premature optimization collapse.

We begin by generalizing prior text-only single-rollout RLVR formulations \cite{xu2025single, brantley2025accelerating} to the multimodal domain. Specifically, for each multimodal input, we generate a single rollout, model the binary reward as a Bernoulli random variable, and estimate the baseline for advantage calculation using a Beta distribution~\citep{johnson1995continuous, mackay2003information}. We further apply batch-wise normalization to reduce variance in advantage estimation. However, this setup remains unstable, suffering from degraded reasoning accuracy as training progresses, as shown for MVSR in Figure~\ref{fig:page1}.

To address this, we identify a direct and effective remedy: \textbf{entropy-based advantage shaping}. This mechanism regularizes advantage magnitudes according to model output entropy, encouraging balanced exploration and stabilizing multimodal training, as illustrated for MSSR in Figure~\ref{fig:page1}. Furthermore, our systematic analysis shows that other non-trivial strategies, such as a cross-modal regularization approach leveraging a parallel text-only branch (motivated by text-only robustness \cite{ko2025flexjudge}) or several regularization techniques (e.g., KL regularization with reference policy, entropy loss \cite{wang2025perception, cheng2025reasoning}), are insufficient to prevent optimization collapse. While previous studies \cite{cheng2025reasoning, liu2025vogue, cui2025entropy} applied entropy shaping within group-based RLVR, we demonstrate its essential and previously underexplored role in stabilizing the multimodal single-rollout setting. In this context, where the lack of intra-group normalization amplifies instability, we show it is not merely beneficial but empirically essential. 

To validate our approach, we conduct experiments using Qwen2.5-VL-3B and 7B models~\citep{Qwen-VL} trained on the Vision-R1-RL dataset~\citep{huang2025vision}. Our results show that MSSR effectively resolves the training efficiency-stability trade-off. First, MSSR is substantially more compute-efficient and effective than the strong group-based GRPO baseline. In in-distribution evaluations, MSSR shows superior training compute efficiency: it achieves GRPO's final validation accuracy with half the total training steps (see Figure~\ref{fig:page1}). Second, MSSR-trained policies exhibit stronger generalization. We evaluate performance across five diverse mathematical and general-domain multimodal reasoning benchmarks: MathVerse \citep{zhang2024mathverse}, MathVista \citep{lu2023mathvista}, MMK12 \cite{meng2025mm}, R1-Onevision-Bench \cite{yang2025r1}, and HallusionBench \citep{guan2024hallusionbench}. On these benchmarks, MSSR shows consistent improvements and outperforms representative group-free and group-based baselines at both model scales, boosting accuracy by an average of 2.1\%  and 2.3\% for 3B and 7B models, respectively (Table~\ref{tab:main_1}).

In summary, our contributions are as follows:

\begin{itemize}
\item We present a systematic study of \textbf{single-rollout} RLVR for multimodal reasoning, identifying key factors that hinder stability compared to text-only settings.
\item We introduce \textbf{MSSR}, a stable and compute-efficient single-rollout RLVR method that matches or surpasses group-based baselines across five benchmarks while greatly improving the training compute efficiency.
\item We demonstrate through extensive ablations that \textbf{entropy-based advantage shaping} is the most effective stabilization strategy among several strong alternatives.
\end{itemize}

% To address these challenges, we present \textbf{MSSR} (\textbf{M}ultimodal \textbf{S}tabilized \textbf{S}ingle-\textbf{R}ollout), a group-free approach that requires only one rollout per multimodal input, aiming to achieve both high sample efficiency and stable optimization. While single-rollout RLVR has recently been studied in text-only settings \cite{xu2025single, brantley2025accelerating}, we find that it suffers from severe instability when extended to multimodal reasoning. Incorporating the image modality introduces substantially higher input variance and significantly complicates cross-modal credit assignment \cite{liu2025vogue, feng2025rewardmap}. Our experiments verify this limitation: naive single-rollout methods that work for text-only RLVR fail to transfer effectively, frequently resulting in premature training collapse. 

%% file: sec/related_updated.tex
\section{Related Work}

\paragraph{Multimodal RLVR.}
Reinforcement Learning with Verifiable Rewards (RLVR) has proven highly effective in enhancing the reasoning capabilities of LLMs \cite{luong2024reft, lambert2024t, guo2025deepseek, zheng2025parallel, dai2025cde}. Recently, the multimodal community has extended this paradigm to MLLMs, showing that verifiable reward signals can improve reasoning across both visual and textual modalities \cite{liu2025vogue, li2025self, huang2025vision, wang2025vl, deng2025openvlthinker, meng2025mm, chen2025sft, wang2025perception}.

These efforts have explored diverse strategies to improve MLLM reasoning, such as using vision-grounded prompts~\cite{huang2025vision}, extending textual reasoning to visual inputs~\cite{yang2025r1}, unifying visual and textual signal within the policy framework~\cite{chen2025sft}. Some approach encourages models to generate visually grounded responses and employ a double-entropy loss to mitigate the training instability caused by a two-branch design \cite{wang2025perception}. Subsequent work has explored more complex designs. For example, \citet{li2025self} developed a framework that decomposes reasoning into visual perception and language inference to improve visual grounding and reduce hallucination, while \citet{meng2025mm} proposed hierarchical visual abstractions for RL-guided planning, and \citet{liu2025vogue} used visual uncertainty to guide exploration for MLLMs. Other studies focus on improving generalization such as leveraging large-scale visual instruction tuning for better cross-modal generalization~\cite{deng2025openvlthinker} and refined reasoning via iterative visual reflection~\cite{wang2025vl}.

However, current multimodal RLVR methods face notable limitations. Most existing approaches rely on group-based optimization (e.g., GRPO), which, while effective, introduces compute-efficiency challenges. Each input sample requires generating multiple responses, leading to repeated forward passes through both the vision and language encoders, which becomes particularly burdensome for large multimodal models. Moreover, when all rollouts within a group yield identical outcomes (e.g., all correct or all incorrect), these methods can encounter reduced rollout utilization efficiency \cite{yu2025dapo}, motivating exploration of more stable and efficient alternatives. In contrast, our method MSSR keeps only one trajectory per input and replaces groupwise normalization with a conjugate Beta baseline for Bernoulli rewards, plus entropy-based advantage shaping. This design directly addresses both compute (no groups) and stability (preventing entropy collapse), enabling efficient yet stable multimodal RLVR at scale.

\vspace{-10pt}
\paragraph{Rollout Efficiency in RLVR.}
A growing line of RLVR research investigates how many rollouts to generate per input and which subset to use for updates to improve efficiency. This research primarily investigates more efficient sampling strategies. For example, some methods filter uninformative prompts before the rollout by filtering zero-variance prompts to avoid uninformative sampling~\cite{zheng2025act}, while others generate many candidates but down-sample them after the rollout to obtain an informative subset~\cite{xu2025not} for reducing redundant computation for each prompt. Another line of research focuses on adaptive, multi-stage rollouts that expand the candidate set only as needed~\cite{zhang2025improving} and using speculative rollouts to accelerate policy updates~\cite{liu2025spec}, which trades modest inference overhead for faster policy updates. In parallel, recent studies in text-only reasoning \cite{xu2025single, brantley2025accelerating} explored single-rollout optimization, replacing group-based comparisons with a single trajectory per prompt.

However, most prior studies on RLVR efficiency have been limited to text-only settings. We find that extending single-rollout optimization to multimodal reasoning is highly non-trivial, introducing instability due to the high-dimensional visual input variance and cross-modal credit assignment~\cite{feng2025rewardmap, liu2025vogue}. Strategies such as batch normalization~\cite{xu2025single, hu2025reinforce++} that work for text-only RLVR are less effective in this setting. To address these issues, we propose MSSR, to the best of our knowledge, the first approach designed to make multimodal RLVR both compute-efficient and training-stable.

%  Visual inputs add high-dimensional variance and cross-modal credit assignment , which amplify training instability and make naive baselines brittle; batch normalization heuristics  that may suffice in text-only reasoning are observed to struggle maintaining stable training in multimodal settings. 
% https://arxiv.org/pdf/2509.22638

% remax （https://arxiv.org/abs/2310.10505）
% rloo (https://arxiv.org/abs/2402.14740)
% spo: https://arxiv.org/pdf/2509.13232

%% file: sec/prelim.tex
\section{Preliminaries}
\paragraph{Group Relative Policy Optimization.}
In the multimodal group-based approach GRPO \citep{shao2024deepseekmath}, given an input $x = (x_\text{text}, x_\text{image})$, a group of responses $\{o_i\}_{i=1}^G$ are sampled from the old policy $\pi_{\theta_\text{old}}$, each associated with a reward $r_i$. Then the normalized advantage for response $o_i$ is defined as $A_i = \frac{r_i - \text{mean}(\{r_i\}_{i=1}^G)}{\text{std}(\{r_i\}_{i=1}^G)}$.

GRPO uses clipped importance sampling to stabilize policy updates.  
Let $\rho_i(\theta) = \frac{\pi_\theta(o_i \mid x)}{\pi_{\theta_{\text{old}}}(o_i \mid x)}$ denote the probability ratio between the new and old policies. The GRPO objective is to maximize the following: 
\vspace{-8pt}
\begin{align}
\label{eq:grpo}
\mathcal{J}_{\text{GRPO}}(\theta)
= &\mathbb{E}_{x \sim \mathcal{D},\, \{o_i\} \sim \pi_{\theta_{\text{old}}}(\cdot \mid x)} \Bigg[
    \frac{1}{G} \sum_{i=1}^G 
    \min\!\big(
        \rho_i(\theta) A_i, \nonumber \\
        & \text{clip}\!\big(\rho_i(\theta), 1\!-\!\epsilon, 1\!+\!\epsilon\big) A_i
    \big)
\Bigg], 
\end{align}
\vspace{-4pt} 
where $\epsilon$ is the clipping hyperparameter.

%% file: sec/approach.tex
\section{Approach}

\begin{figure*}[!t]
    \centering
    \includegraphics[width=0.9\textwidth]{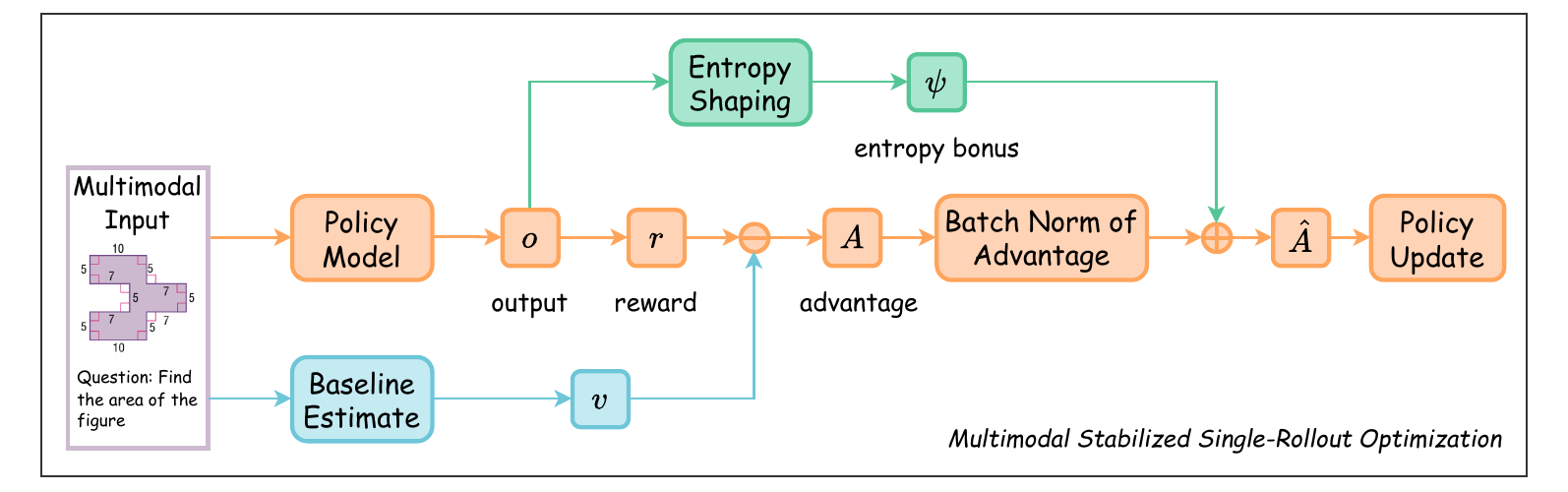}
    \caption{\textbf{Overview of the proposed MSSR approach.} Given a multimodal input,~\ie, an image and the corresponding question, we generate a single rollout through the policy model. We then use a Beta distribution to estimate the baseline value $v$, compute the advantage $A$, and normalize it across the batch. Finally, we propose entropy-based advantage shaping to preserve entropy and stabilize training.}
    \label{fig:approach}
    \vspace{-10pt}
\end{figure*}

Our approach aims to achieve efficient yet stable multimodal RLVR, addressing the rollout utilization inefficiency of group-based methods, which becomes particularly burdensome for large multimodal models. To this end, we first propose a multimodal vanilla single-rollout approach that generates only one trajectory per input and replaces groupwise normalization with a conjugate Beta baseline for binary Bernoulli rewards. However, we observe entropy collapse and training instability, as the lack of intra-group normalization amplifies variance and instability. To overcome this issue, we incorporate an entropy-based advantage shaping mechanism to preserve model entropy and stabilize training, which yields our Multimodal Stabilized Single-Rollout (MSSR) approach, as shown in Figure \ref{fig:approach}.

\vspace{-10pt}
\paragraph{Problem Formulation.}
We consider a multimodal reasoning task where the input $x = (x_\text{text}, x_\text{image})$ consists of a textual prompt $x_\text{text}$ and an associated image $x_\text{image}$. The multimodal policy model $\pi_\theta(o|x)$ generates a single-rollout output $o$. A verifiable reward function $r(x, o) \in \{0, 1\}$ evaluates the correctness of the generated answer. The objective of our multimodal RLVR approach is to maximize the expected verifiable reward under the model distribution: $\mathcal{J}(\theta) = \mathbb{E}_{x \sim \mathcal{D},\, o \sim \pi_\theta(\cdot|x)}[\,r(x, o)\,]$.

\vspace{-10pt}
\paragraph{Multimodal Vanilla Single-Rollout.} \label{sec:mvsr}
Since the reward $r(x, o)$ is binary, it can be modeled as a random variable with Bernoulli distribution. The baseline estimate of the expected reward naturally follows a Beta distribution \cite{johnson1995continuous, mackay2003information}. Therefore, for each multimodal input $x$, we maintain a Beta distribution \(\text{B}(\alpha(x), \beta(x))\) with shape parameters $\alpha$ and $\beta$ to estimate the baseline value for advantage calculation, generalizing beyond prior works for text-only reasoning \citep{xu2025single, xiao2025bnpo}. We compute the baseline value as the mean of the Beta distribution:
$
\hat{v}(x) = \frac{\alpha(x)}{\alpha(x) + \beta(x)}.
$
After observing a reward \(r(x, o)\) at each training step, we update the parameters as
$\alpha(x) \leftarrow \eta \cdot \alpha(x) + r(x, o)$, $\beta(x) \leftarrow \eta \cdot \beta(x) + (1 - r(x, o))$, 
where $\eta \in [\eta_{\min}, \eta_{\max}] \subset (0, 1]$ is an adaptive discount factor 
that varies during training to enable better adaptation as the policy evolves. 
We calculate the advantage as
$
A(x, o) = r(x, o) - \hat{v}_{-1}(x),
$
where $\hat{v}_{-1}(x)$ is the baseline estimate from the previous step 
(to avoid bias \citep{xu2025single}). Then we normalize advantages across the batch. We initialize the baseline estimate $\hat{v}_0(x)$ as the reward $r(x, o)$ obtained by a forward pass using the initial policy $\pi_0$ before training begins. We then set the initial parameters of the Beta distribution as $ \alpha_0(x) = \frac{\hat{v}_0(x)}{1 - \eta_{\min}}$, $\beta_0(x) = \frac{1 - \hat{v}_0(x)}{1 - \eta_{\min}}$, similar to prior initialization strategies \cite{xu2025single}. 

To adaptively update the discount factor $\eta$, we maintain a sliding window of size $N$ over training steps to track the KL divergence between the policies of two consecutive updates. We set the initial value of $\eta$ as $\eta_0 = \frac{\eta_{\min} + \eta_{\max}}{2}$. For each step $s$, the average KL divergence over the last $N$ steps is denoted as $\widebar{\mathrm{KL}}_s$. If $\widebar{\mathrm{KL}}_s > \mathrm{KL}_{\text{target}}$, where $\mathrm{KL}_{\text{target}}$ is a target KL divergence value, we compute a ratio $\tau_s$ and adjust $\eta_s$ using a linear decay rule:
\begin{align} \label{eq:eta_decay}
\vspace{-5pt}
\tau_s &= \min\left(\frac{\widebar{\mathrm{KL}}_s}{\mathrm{KL}_{\text{target}}},\, 1.0\right), \\ \nonumber
\eta_s &= \eta_{\max} - \tau_s \cdot \left(\eta_{\max} - \eta_{\min}\right).
\vspace{-5pt}
\end{align}
When the KL divergence is high, indicating that the policy is changing rapidly, a smaller $\eta$ enforces faster forgetting, preventing instability during optimization. Conversely, if $\widebar{\mathrm{KL}}_s \leq \mathrm{KL}_{\text{target}}$, we apply a linear growth rule:
\begin{align} \label{eq:eta_grow}
\tau_s = \frac{\widebar{\mathrm{KL}}_s}{\mathrm{KL}_{\text{target}}}, \quad
\eta_s = \eta_{\min} + \tau_s \cdot \left(\eta_{\max} - \eta_{\min}\right).
\end{align}

When the KL divergence is low, suggesting that the policy updates are stable, a larger $\eta$ results in slower forgetting, allowing the model to retain more information from previous updates. 

By performing policy gradient optimization with the obtained advantage, we obtain Multimodal Vanilla Single-Rollout (\textbf{MVSR}) as the baseline approach.

% \begin{figure*}[t]
%     \centering
%     \includegraphics[width=\textwidth]{figs/case2.png}
%     \caption{\textbf{Example of reasoning outputs.} comparing the Qwen2.5-VL-7B model fine-tuned with GRPO, MVSR, and MSSR. While GRPO and MVSR produce incorrect answers, MSSR successfully solves the problem, demonstrating its superior reasoning capability.}
%     \label{fig:case}
%     \vspace{-10pt}
% \end{figure*}

\vspace{-8pt}
\paragraph{Multimodal Stabilized Single-Rollout (MSSR).} \label{sec:mssr}
In the multimodal single-rollout RLVR setting, because only a single response is sampled per input, the observed reward signal could be highly stochastic: a response that happens to yield $r=1$ may be followed by another response yielding $r=0$, even for similar reasoning trajectories. This high variance can make the policy update volatile, particularly in multimodal reasoning tasks where the visual component introduces additional uncertainty. Consequently, direct advantage estimation from binary rewards can lead to \emph{entropy collapse} and further unstable training. And this hypothesis is verified by the results in Figure \ref{fig:entropy}, where the entropy of MVSR collapses and resulting in degraded accuracy. 

To address this instability of vanilla single-rollout approach, we incorporate an \emph{entropy bonus} $\psi$ that integrates the policy's entropy directly into the advantage. This bonus is based on the token entropy $\mathcal{H}_t(\pi_\theta)$ of the policy's output distribution, which is defined as $\mathcal{H}_t(\pi_\theta) = - \mathbb{E}_{o \sim \pi_\theta(\cdot|x)} [ \log \pi_\theta(o_{<t}|x)]$. We then define the entropy bonus as $\psi_t = \min\!\Big(\frac{|A_t|}{\gamma}, \lambda \cdot \mathrm{stopgrad}(\mathcal{H}_t) \Big)$, where $\gamma$ and $\lambda$ are scaling factors, $\mathrm{stopgrad(\cdot)}$ is the stop gradient operator. Therefore, for our multimodal stabilized single-rollout (\textbf{MSSR}) policy, we 
propose the \textbf{entropy-based advantage shaping} by computing the shaped advantage as $\hat{A}_t = A_t + \psi_t$.

Intuitively, this modification encourages the policy to assign higher effective advantage to responses generated under higher uncertainty,
even if their reward is low, since such responses may explore promising but under-sampled reasoning trajectories. It reshapes the advantage by softening the penalty for low-reward responses that may still lie near correct reasoning trajectories, and it maintains sufficient policy entropy to promote exploration and mitigate mode collapse.

%% file: sec/experiments.tex
\vspace{-5pt}
\section{Experiments} \label{sec: exp}

\subsection{Experimental Setup} \label{sec: setup}

\paragraph{Implementation Details.}
We perform direct RL training on top of Qwen2.5-VL-3B and 7B \citep{Qwen-VL} models. The models are trained to produce structured outputs, where the reasoning process is enclosed in \texttt{<think></think>} and the final answer is formatted using \texttt{\textbackslash boxed\{\}}.

We train on the Vision-R1-RL dataset~\citep{huang2025vision}, which contains approximately $10$K samples designed to be suitable for RLVR. This dataset focuses on the multimodal math reasoning domain, using real-world images (e.g., charts, diagrams, and visual math problems). The verifiable rewards are computed using a hard formatting rule-based binary reward function, which provides a positive reward only if the final answer exactly matches the ground-truth solution.  All models are trained for $120$ steps on $8$ GPUs using AdamW~\citep{loshchilov2018decoupled} with a learning rate of $1\times10^{-6}$ and weight decay of $0.01$. For entropy-based advantage shaping, we set $\gamma = 0.4$ and $\lambda = 2.0$, following the setup of \cite{cheng2025reasoning}. The discount factor range for updating the Beta distribution parameters is set to $\eta_{\min} = 0.875$ and $\eta_{\max} = 0.96$, consistent with \cite{xu2025single}. We use a sliding window of size $N = 20$ to track the KL divergence between consecutive policy updates and set the target KL value to $\mathrm{KL}_{\text{target}} = 0.01$ for adjusting $\eta$. The design choices for these hyperparameters are analyzed in Section~\ref{sec:sensitivity}. We apply KL regularization to the reference policy with a coefficient of $0.01$.  The implementations are built on the EasyR1 framework \citep{zheng2025easyr1}.

\begin{table*}[ht]
    \centering
    \caption{\textbf{Model generalization performance on diverse multimodal reasoning benchmarks.} We compare MSSR with GRPO, RLOO, and REINFORCE++ baselines on Qwen2.5-VL 3B and 7B models. For broader context, we also report evaluated results from prior SFT+RL and Zero-RL methods following the evaluation protocol of \cite{zhu2025shuffle}. MSSR outperforms other baselines, with Qwen2.5-VL-7B + MSSR achieving the strongest average performance across benchmarks.}
    \vspace{-5pt}
    \resizebox{\textwidth}{!}{
    \begin{tabular}{lcccccc}
        \toprule
        \textbf{Model} & \textbf{MathVerse} & \textbf{MathVista}  & \textbf{MMK12} & \textbf{\makecell{R1-Onevision \\ Bench}} & \textbf{HallusionBench}  & \textbf{Avg.}  \\
        \midrule
        \multicolumn{7}{c}{\text{SFT + RL}} \\ 
        \midrule
        R1-Onevision-7B \citep{yang2025r1} & 46.0 & 62.9  & 43.5 & 35.2 & 67.2  & 51.0 \\
        OpenVLThinker-7B \citep{deng2025openvlthinker} & 45.8 & 70.0  & 53.5 & 34.7 & 60.0 & 52.8 \\
        % Vision-R1-7B \citep{huang2025vision} & 50.2 & 71.2   & 47.5 & 33.5 & 57.8  & 52.0 \\
        VLAA-Thinker-7B \citep{chen2025sft} & 48.2 & 68.0  & 51.7 & 38.4 & 70.0  & 55.3 \\
        \midrule
        \multicolumn{7}{c}{\text{Zero RL}} \\
        \midrule
        MM-Eureka-Qwen-7B \citep{meng2025mm} & 50.3  & 71.2  & 61.7 & 39.1 & 66.4 & 57.7 \\
        ThinkLite-VL-7B \citep{wang2025sota} & 47.3 & 71.9    & 57.6 & 35.7 & 70.9  & 56.7 \\
        % VL-Rethinker-7B \citep{wang2025vl} & 73.3 & 49.5  & 62.4 & 43.9 & 69.5  & 59.7 \\
        \midrule
        Qwen2.5-VL-3B \citep{Qwen-VL} & 33.3 & 59.5   & 42.5 & 27.6 & 59.9  & 44.6 \\
        \quad + GRPO \cite{shao2024deepseekmath} & 36.8 & 61.7    & 46.1 & \textbf{30.2} & 62.3  & 47.4 \\ 
        \quad + RLOO \cite{ahmadian2024back} & 35.7 & 59.7   & 45.5 & 28.8 & 61.6 & 46.3  \\ 
        \quad + REINFORCE++ \cite{hu2025reinforce++} & 35.3 & 47.7   & 46.0 & 21.7 & 63.2  & 42.8  \\ 
        \quad + MSSR  & \textbf{39.6} & \textbf{63.0}  & \textbf{49.2} & 29.0 & \textbf{66.6}  & \textbf{49.5} \\
        \midrule
        Qwen2.5-VL-7B \citep{Qwen-VL} & 45.8 & 67.2   & 48.1 & 34.6 & 68.4 & 52.8 \\
        \quad + GRPO \cite{shao2024deepseekmath} & 48.5 & 70.0   & 55.8 & 37.7 & 69.7  & 56.3  \\ 
        \quad + RLOO  \cite{ahmadian2024back} & 47.8 & 69.2   & 56.0 & 38.5 & 68.5  & 56.0  \\ 
        \quad + REINFORCE++ \cite{hu2025reinforce++} & 42.7 & 68.5    & 51.3 & 34.0 & 69.2  & 53.1 \\
        \quad + MSSR & \textbf{49.8} & \textbf{71.1}   & \textbf{62.5} & \textbf{39.2} & \textbf{70.6}  & \textbf{58.6}  \\
        \bottomrule
    \end{tabular}
    }
    \label{tab:main_1}
    \vspace{-10pt}
\end{table*}

\vspace{-10pt}
\paragraph{Baselines.}
We compare MSSR to both group-based and group-free baselines under the same setup. For the group-based baselines, GRPO~\citep{shao2024deepseekmath} and RLOO~\citep{ahmadian2024back}, we adopt a batch size of $256 \times 8 = 2048$: $256$ unique prompts and $n = 8$ rollouts generated per prompt. For the group-free setting (REINFORCE++ \citep{hu2025reinforce++} with single-rollout) and our proposed MSSR, we ensure fairness by using an equivalent batch size of $2048 \times 1 = 2048$. For broader context, we also report evaluated results from prior 7B models: R1-Onevision-7B \citep{yang2025r1}, OpenVLThinker-7B \citep{deng2025openvlthinker}, VLAA-Thinker-7B \citep{chen2025sft}, MM-Eureka-Qwen-7B \citep{meng2025mm}, and ThinkLite-VL-7B \citep{wang2025sota}.

% and VL-Rethinker-7B \citep{wang2025vl}.

\vspace{-6pt}
\paragraph{Evaluation Benchmarks.}
We evaluate on five multimodal reasoning benchmarks for model generalization performance, including MathVerse \citep{zhang2024mathverse}, MathVista \citep{lu2023mathvista}, MMK12 \cite{meng2025mm}, R1-Onevision-Bench \cite{yang2025r1}, HallusionBench \citep{guan2024hallusionbench}. These benchmarks span diverse aspects of multimodal reasoning, covering mathematical problem solving, multidisciplinary reasoning (math, physics, chemistry, biology), and hallucination detection. For evaluation, we follow the protocol of \cite{zhu2025shuffle} and use Qwen2.5-72B-Instruct~\citep{qwen2.5} to extract final answers from model responses~\citep{su2025crossing}, and we measure correctness by matching them against reference answers following prior work~\citep{liu2025vogue}.

%  and LogicVista \citep{xiao2024logicvista}

\subsection{Main Results}
\begin{figure}[t]
    \centering
    \includegraphics[width=0.85\linewidth]{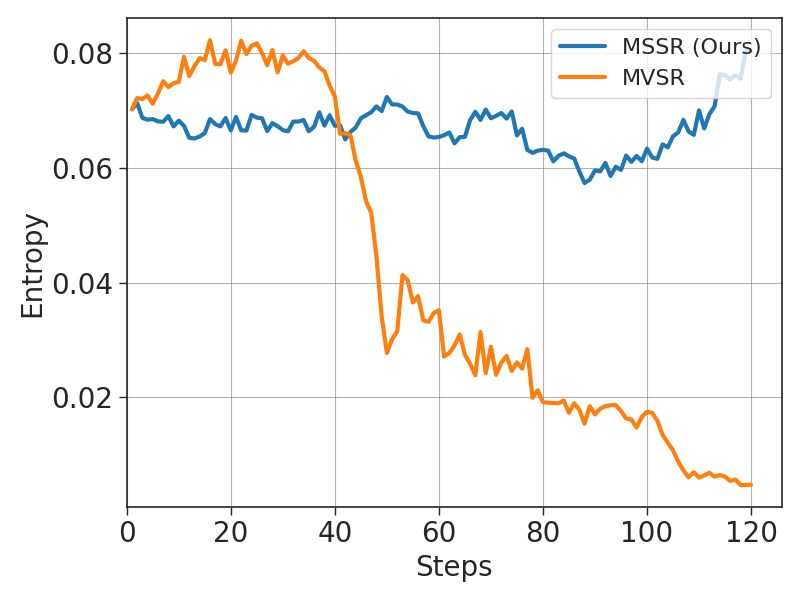}
    \vspace{-5pt}
    \caption{\textbf{Model output entropy during training with Qwen2.5-VL-7B.} MVSR (multimodal vanilla single-rollout) suffers from entropy collapse as training progresses, whereas our proposed MSSR (multimodal stabilized single-rollout) preserves entropy.}
    \label{fig:entropy}
    \vspace{-10pt}
\end{figure}

From Figures \ref{fig:page1}\ref{fig:train_acc} and \ref{fig:page1}\ref{fig:val_acc}, we observe that the Multimodal Vanilla Single-Rollout (MVSR) approach is unstable, with both training and validation accuracy degrading as training progresses. As shown in Figure \ref{fig:entropy}, this degradation in MVSR performance correlates with entropy collapse. To address this, we incorporate entropy-based advantage shaping as described in Section \ref{sec:mssr}, which regularizes the advantage magnitude according to the model entropy. As illustrated by the MSSR (Multimodal Stabilized Single-Rollout) curve in Figure~\ref{fig:entropy}, this prevents entropy collapse, and the resulting MSSR approach achieves stable training dynamics, with both training and validation accuracy improving steadily (Figures \ref{fig:page1}\ref{fig:train_acc} and \ref{fig:page1}\ref{fig:val_acc}). 

We then compare MSSR to the widely used group-based multimodal approach GRPO. Note that for both single-rollout and group-based methods, we use an equivalent total number of rollouts per step (2048), although with different configurations (1 and 8 rollouts per input, respectively). We first measure training cost as the average training time per step (mins/step) across methods. See the detailed results in Table~\ref{tab:time} in Appendix. Although MSSR incurs minimal average per-step overhead (6.9 vs. 6.1 mins/step due to baseline estimation), it achieves similar validation accuracy to GRPO with half of the total training steps, demonstrating substantially higher rollout utilization efficiency. When trained for a comparable computational budget, MSSR further surpasses GRPO in final performance (Figures \ref{fig:page1}\ref{fig:train_acc} and \ref{fig:page1}\ref{fig:val_acc}). 

%MSSR also yields higher validation accuracy, indicating better reasoning capability.

% GRPO requires 6.1 mins/step, while MSSR requires 6.9 mins/step. The minimum overhead in MSSR arises from an initial forward pass to estimate baseline values. Under comparable number of steps, MSSR achieves faster training progress and reaches a higher final training accuracy . Notably, MSSR attains accuracy comparable to GRPO in less than half the number of training steps, demonstrating superior sample efficiency. 

Furthermore, we evaluate model generalization performance on a diverse set of multimodal reasoning benchmarks, including MathVerse, MathVista, MMK12, R1-OneVision-Bench, and HallusionBench (Table~\ref{tab:main_1}). We compare MSSR against GRPO, RLOO, and REINFORCE++ in terms of accuracy on Qwen2.5-VL 3B and 7B backbones. Under comparable training budgets, MSSR consistently improves average performance across benchmarks, demonstrating enhanced multimodal reasoning. And Qwen2.5-VL-7B + MSSR achieves the strongest mean performance among all evaluated methods.

% \begin{table*}[ht]
%     \centering
%     \caption{\textbf{Model performance of pass@4 accuracy on diverse visual reasoning benchmarks.} On both Qwen2.5-VL 3B and 7B models, VOGUE consistently improves over GRPO and achieves the highest average pass@4 accuracy.}
%     % \resizebox{\textwidth}{!}{
%     \begin{tabular}{lccccccc}
%         \toprule
%         \textbf{Model} & \textbf{MathVerse} & \textbf{MathVista} & \textbf{MMK12} & \textbf{\makecell{R1-Onevision \\ Bench}} & \textbf{HallusionBench} & \textbf{LogicVista} & \textbf{Avg.}\\
%         \midrule
%         Qwen2.5-VL-3B \citep{Qwen-VL} & 55.5 & 78.0 & 73.1 & 55.4 & 83.1 & 72.7 \\
%         \quad + GRPO \cite{shao2024deepseekmath} & 56.1 & 78.0 & 75.3 & 55.1 & 84.2 & 71.4 \\ 
%         \quad + RLOO \cite{ahmadian2024back} & 56.2 & 78.2 & 75.1 & 55.9 & 83.3 & 72.3 \\ 
%         \quad + REINFORCE++ \cite{hu2025reinforce++} & 48.7 & 61.6 & 67.2 & 36.0 & 76.5 & 42.6 \\ 
%         \quad + MSSR & 55.5 & 78.1  & 76.4 & 52.0 & 84.2 & 65.4 \\
%         \midrule
%         Qwen2.5-VL-7B \citep{Qwen-VL} & 60.2 & 82.1 & 74.2 & 59.3 & 85.1 & 72.3 \\
%         \quad + GRPO \cite{shao2024deepseekmath} & 62.2 & 84.1 & 79.5 & 61.0 & 82.0  & 74.5 \\ 
%         \quad + RLOO \cite{ahmadian2024back} & 61.9 & 83.1 & 81.1 & 63.1 & 82.3 & 74.5 \\ 
%         \quad + REINFORCE++ \cite{hu2025reinforce++} &  &  &  &  &  &  \\ 
%         \quad + MSSR & 58.7 & 81.4 & 80.3 & 58.6 & 81.4 & 73.7 \\
%         \bottomrule
%     \end{tabular}
%     % }
%     \label{tab:main_3}

% \end{table*}

\subsection{Ablation Studies}
For the ablation studies, we explore several techniques to assess their effectiveness in stabilizing training or preventing entropy collapse for the multimodal single-rollout optimization, including KL regularization with reference policy, cross-modal regularization, and entropy loss.

\paragraph{KL Regularization with Reference Policy.}
KL regularization is a commonly used mechanism to constrain policy updates and prevent the policy from drifting excessively during optimization for stability. While we applied it in our experiments for such purpose, we find it insufficient on its own to stabilize training in the multimodal single-rollout setting. This is clearly demonstrated by the entropy collapse of the multimodal vanilla single-rollout (MVSR) approach in Figure~\ref{fig:entropy} and performance drop shown in Figures~\ref{fig:page1}\ref{fig:train_acc} and~\ref{fig:page1}\ref{fig:val_acc}.  The training still diverges and accuracy degrades, indicating that additional stabilization mechanisms are required for multimodal single-rollout RLVR.

% This indicates that KL regularization is insufficient to prevent instability in the multimodal single-rollout regime.

\begin{figure*}[t]
\vspace{-0.2in}
    \centering
    \begin{subfigure}[t]{0.33\linewidth}
        \centering
        % \captionsetup{justification=raggedright,singlelinecheck=false}
        \includegraphics[width=\textwidth]{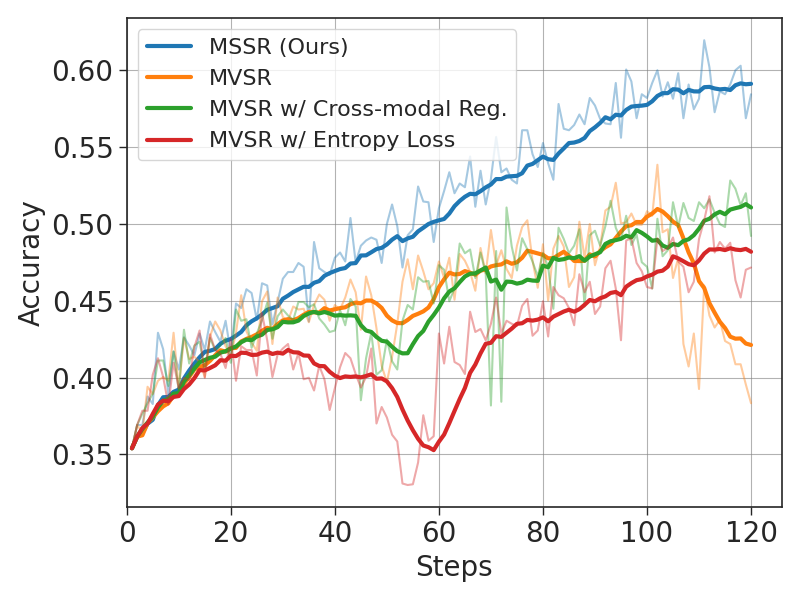}
        \caption{Training accuracy}
        \label{fig:abl_train}
    \end{subfigure}
    \begin{subfigure}[t]{0.33\linewidth}
        \centering
        \includegraphics[width=\textwidth]{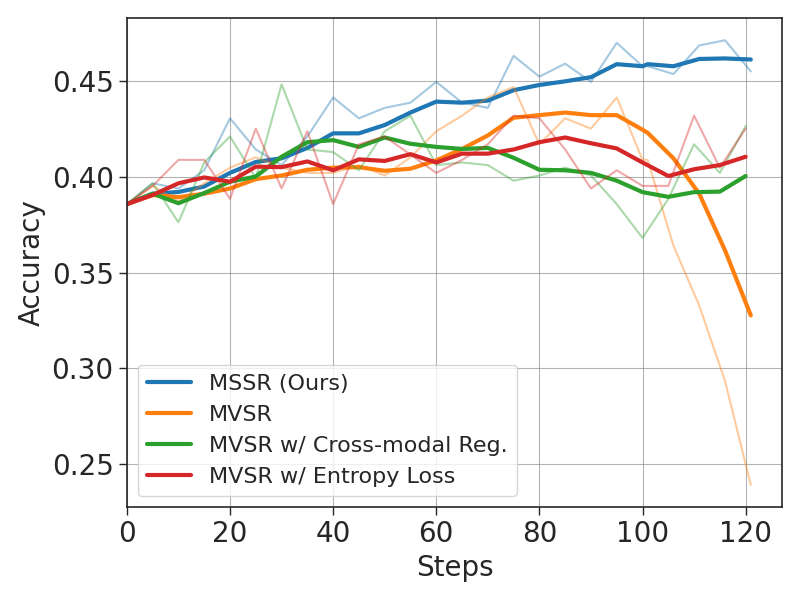}
        \caption{Validation accuracy}
        \label{fig:abl_val}
    \end{subfigure}
    \begin{subfigure}[t]{0.33\linewidth}
        \centering
        \includegraphics[width=\textwidth]{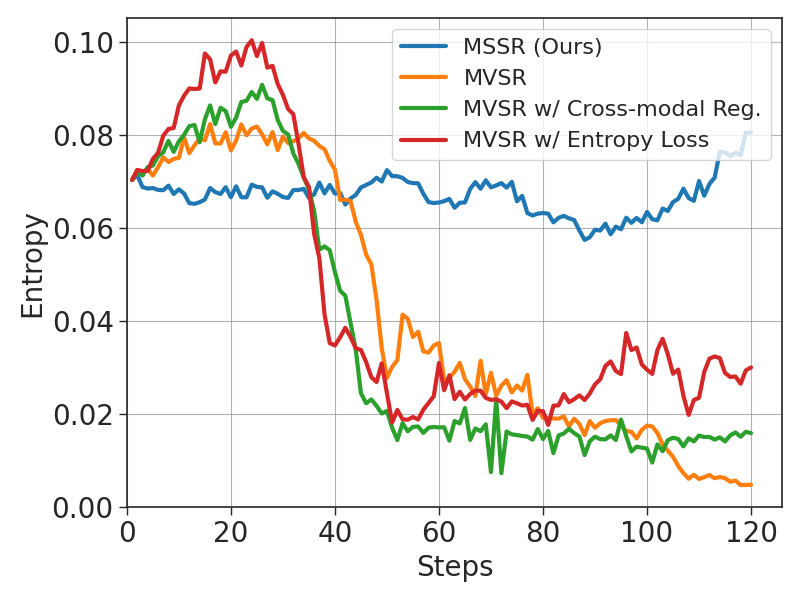}
        \caption{Model entropy}
        \label{fig:abl_entropy}
    \end{subfigure}
    \vspace{-0.1in}
    % \begin{subfigure}[t]{0.33\linewidth}
    %     \centering
    %     % \setlength{\fboxsep}{0pt}
    %     % \fbox{\rule{0pt}{4cm}\rule{\linewidth}{0pt}} % placeholder box
    %     \includegraphics[width=\textwidth]{figs/entropy_loss_reward_accuracy_7b.png}
    %     \caption{Training accuracy}
    %     \label{fig:ent_loss_train}
    % \end{subfigure}
    % \begin{subfigure}[t]{0.33\linewidth}
    %     \centering
    %     \includegraphics[width=\textwidth]{figs/entropy_loss_val_pass@1_7b.png}
    %     \caption{Validation accuracy}
    %     \label{fig:ent_loss_val}
    % \end{subfigure}
    % \begin{subfigure}[t]{0.33\linewidth}
    %     \centering
    %     \includegraphics[width=\textwidth]{figs/entropy_loss_entropy_7b.png}
    %     \caption{Model entropy}
    %     \label{fig:ent_loss_entropy}
    % \end{subfigure}

\caption{\textbf{Ablation studies on effectiveness of techniques for preventing entropy collapse and stabilizing multimodal single-rollout training.} \textbf{Cross-modal regularization:} This technique provides partial stabilization, increasing training accuracy but still resulting in degraded validation accuracy, and both metrics remain below those achieved by MSSR. \textbf{Entropy loss:} Adding an entropy loss term partially preserves entropy and improves training accuracy toward the end of training, but validation performance still degrades and entropy is not maintained as effectively as in MSSR.}
    \label{fig:ablation}
    \vspace{-10pt}
\end{figure*}

\vspace{-6pt}
\paragraph{Cross-Modal Regularization.}
The multimodal policy optimization process can be unstable, whereas text-only updates are typically more robust \cite{ko2025flexjudge}. Therefore, we examine another non-trivial regularization approach, which is to use cross-modal regularization that incorporates a text-only branch alongside the multimodal branch. The text-only branch can acts as a stabilizing “anchor policy” that might keep reasoning grounded and prevents mode collapse. Specifically, both branches share the same textual prompt as input, while the image input is omitted for the text-only branch. The text-only branch uses the same policy model as the multimodal branch but does not undergo policy updates with verifiable rewards; instead, it serves solely as an anchor policy that provides a reference for KL divergence regularization against the multimodal policy. We then measure the KL divergence between their output distributions as: 
$\mathcal{L}_{\text{KL}} = \mathbb{E}
\big[
\mathrm{KL} \big( \pi_\theta(\cdot \mid x_\text{text}, x_\text{image}) \;\Vert\; \pi_\theta(\cdot \mid x_\text{text}) \big)
\big]$.

We incorporate this KL divergence term as a regularization component into the multimodal policy loss instead of using the KL divergence to the fixed reference policy, aiming to encourage the multimodal policy to stay aligned with its more stable text-only counterpart during optimization and thereby improving training stability.

However, as shown in Figures \ref{fig:abl_train}, \ref{fig:abl_val}, and \ref{fig:abl_entropy}, this technique provides only partial stabilization: training accuracy improves, but validation accuracy still degrades, and both metrics remain below those achieved by MSSR.

\vspace{-10pt}
\paragraph{Entropy loss.}
We next examine the effects of adding an entropy loss term to the policy gradient loss for preventing model entropy collapse. However, as shown in Figures~\ref{fig:abl_train}, \ref{fig:abl_val}, and \ref{fig:abl_entropy}, this approach does not prevent entropy collapse in the multimodal vanilla single-rollout (MVSR) setting. While adding an entropy loss helps preserve entropy to some extent and leads to improved training accuracy toward the end of training, the validation accuracy still degrades, and the entropy is not maintained as effectively as in MSSR. This indicates that using entropy loss is insufficient to stabilize multimodal single-rollout training.

In summary, these ablation studies confirm that our entropy-based advantage shaping mechanism is essential for stabilizing multimodal single-rollout training. While techniques such as cross-modal regularization, and entropy loss offer partial improvements, they fail to consistently prevent entropy collapse or maintain validation performance. In contrast, our advantage shaping strategy effectively preserves model entropy throughout training, yields stable optimization dynamics, and leads to both higher training and validation accuracy, demonstrating its central role in enabling reliable multimodal single-rollout learning. Across all tests, MSSR improves the final validation accuracy by approximately 5\% over the strongest single-rollout variant.

\begin{figure*}[t]
    \centering
    \vspace{-0.2in}
    \includegraphics[width=0.9\linewidth]{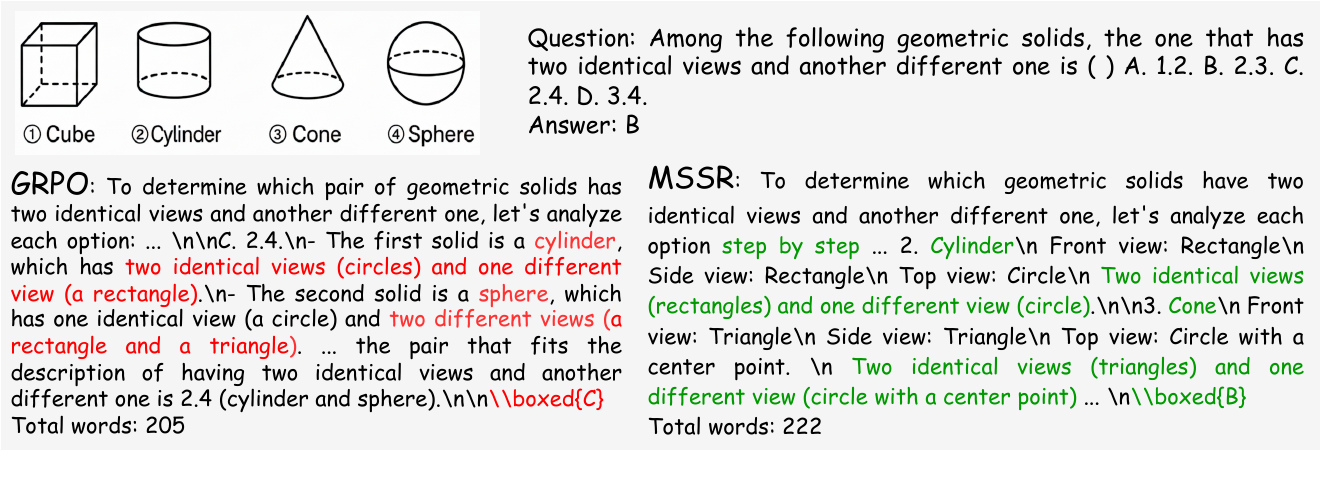}
    \vspace{-20pt}
    \caption{\textbf{Comparison of reasoning outputs from GRPO and MSSR} (Multimodal Stabilized Single-Rollout). MSSR produces the correct answer while GRPO fails. We highlight the critical reasoning steps that lead to GRPO’s incorrect answer in \textcolor{red}{red}, and the key steps enabling MSSR’s correct prediction in \textcolor{ForestGreen}{green}. }
    \label{fig:case1}
    \vspace{-15pt}
\end{figure*}

%  \begin{minipage}{\textwidth}
%   \centering
%   \includegraphics[width=0.9\linewidth]{figs/MLLM_Exploration-Page-5.drawio.pdf}
%   \teasersubcap{Example of reasoning outputs comparing GRPO and MSSR}{fig:case}
% \end{minipage}

\begin{figure}[t]
    \centering
    % \begin{subfigure}[t]{0.49\linewidth}
    %     \centering
    %     \includegraphics[width=\textwidth]{figs/n_reward_accuracy_7b.png}
    %     \caption{Training accuracy}
    %     \label{fig:n_train}
    % \end{subfigure}
    \begin{subfigure}[t]{0.75\linewidth}
        \centering
        \includegraphics[width=\textwidth]{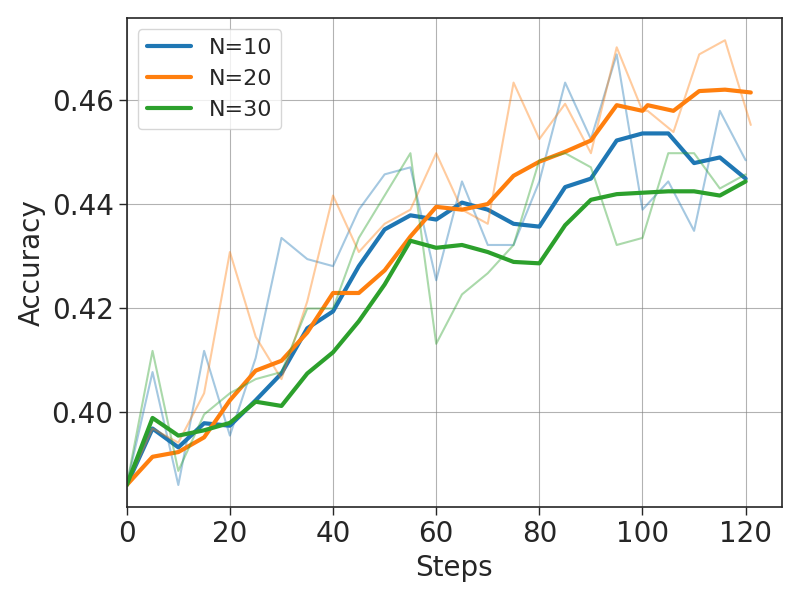}
        \caption{Validation accuracy for varying $N$.}
        \label{fig:n_val}
    \end{subfigure}

    % \begin{subfigure}[t]{0.49\linewidth}
    %     \centering
    %     \includegraphics[width=\textwidth]{figs/kl_reward_accuracy_7b.png}
    %     \caption{Training accuracy}
    %     \label{fig:kl_train}
    % \end{subfigure}
    \begin{subfigure}[t]{0.75\linewidth}
        \centering
        \includegraphics[width=\textwidth]{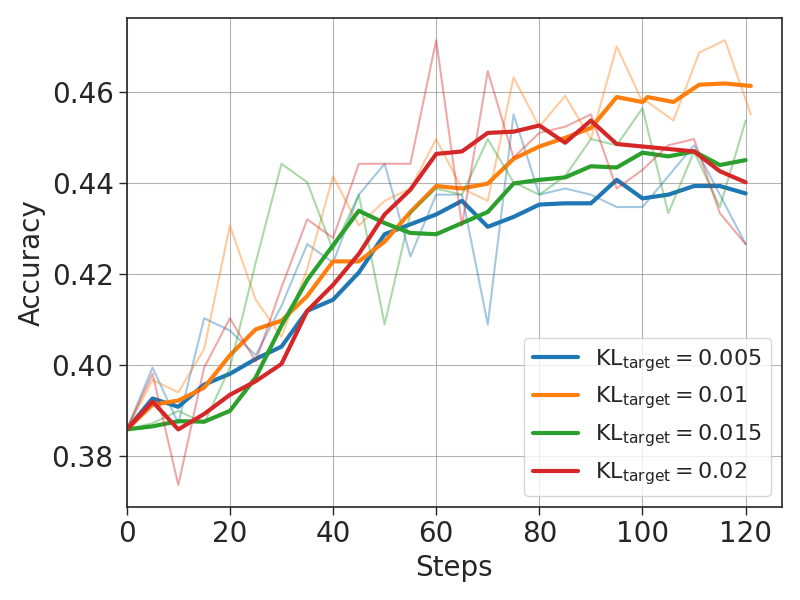}
        \caption{Validation accuracy for varying $\mathrm{KL}_{\text{target}}$.}
        \label{fig:kl_val}
    \end{subfigure}
    \caption{\textbf{Sensitivity analysis for effects of sliding window size $N$ and target KL value $\mathrm{KL}_{\text{target}}$ on adjusting the discount factor $\eta$ and their impact on validation accuracy.} 
    (a) Validation accuracy for different $N$, which determine how the average KL divergence between consecutive policy updates is computed. 
    (b) Validation accuracy for different $\mathrm{KL}_{\text{target}}$, which controls how aggressively $\eta$ adapts. 
    $N=20$ and $\mathrm{KL}_{\text{target}}=0.01$ yield higher validation accuracy at the end of training.}
    \label{fig:sen_val}
    \vspace{-10pt}
\end{figure}

% \vspace{-5pt}
\subsection{Sensitivity Analysis} \label{sec:sensitivity}
We conduct a hyperparameter sensitivity analysis for design choices. We first analyze the effect of the sliding window size $N$, which is used to compute the average KL divergence between consecutive policy updates and thereby adjust the discount factor $\eta$ when updating the Beta distribution parameters (Eq.~\ref{eq:eta_decay} and Eq.~\ref{eq:eta_grow}). We evaluate how different values of $N$ influence training dynamics. As shown in Figure \ref{fig:n_val}, with $N = 20$ yielding higher validation accuracy toward the end of training. Based on this observation, we adopt $N = 20$.

We then examine the effect of the target KL divergence value $\mathrm{KL}_{\text{target}}$, which determines how aggressively $\eta$ is adjusted between updates. As shown in Figure \ref{fig:kl_val}, $\mathrm{KL}_{\text{target}} = 0.01$ achieves higher validation accuracy toward the end of training. Therefore, we set $\mathrm{KL}_{\text{target}} = 0.01$.

The training accuracy curves for different values of $N$ and $\mathrm{KL}_{\text{target}}$ are largely similar; for completeness, we present them in Figure~\ref{fig:sen_train} in Appendix.

% \subsubsection{REINFORCE++ with Entropy Shaping}
% See if group-free REINFORCE++ works with entropy-based advantage shaping for multimodal reasoning. Verify the effects of using Beta distribution to estimate baseline value. 

% \begin{figure}[h]
%     \centering
%     \begin{subfigure}[t]{0.49\linewidth}
%         \centering
%         \includegraphics[width=\textwidth]{figs/cross_modal_val_pass@1_7b.png}
%         \caption{Validation accuracy}
%         \label{fig:cross_val}
%     \end{subfigure}
%     \begin{subfigure}[t]{0.49\linewidth}
%         \centering
%         \includegraphics[width=\textwidth]{figs/cross_modal_entropy_7b.png}
%         \caption{Model entropy}
%         \label{fig:cross_entropy}
%     \end{subfigure}
%     \caption{REINFORCE++ with entropy shaping.}
%     \label{fig:reinforce_shaping}
% \end{figure}

% \subsubsection{Batch-level discount factor vs sample-level discount factor}

% Verify the design choice of why using batch-level discount factor, not sample-level. 

% \begin{figure}[t]
%     \centering
%     % No padding inside the frame
%     \setlength{\fboxsep}{0pt}
%     % Optional: thicker border
%     % \setlength{\fboxrule}{1pt}
%     \fbox{\rule{0pt}{6cm}\rule{\linewidth}{0pt}}
%     \caption{Verify the design choice of why using batch-level discount factor, not sample-level (placeholder)}
%     \label{fig:approach}
% \end{figure}

\subsection{Other Discussions}
We observe that MSSR-trained policies tend to exhibit more \textbf{fine-grained key steps} during problem-solving compared to the GRPO-trained policies. To quantify this observation, we compute the average number of key steps per response. We identify these steps by counting instances of markdown bolding pairs (e.g., **step**), as this formatting is frequently used by the base model to delineate its key reasoning stages. As shown in Table~\ref{tab:key_steps}, MSSR-trained policies produce an average of 3.3 key steps, compared to 1.9 for GRPO and 3.1 for the base model. This quantitative result supports our observation. GRPO appears to converge to shorter solutions, whereas MSSR maintains a more detailed, step-by-step reasoning process that is even slightly more fine-grained than the base model. The fine-grained reasoning solutions may serve as another indicator of a more robust reasoning process and alleviation of model collapse. The trend is also reflected in the average response length, where MSSR's responses are longer than that of GRPO, showing that MSSR produces a more detailed and well-formed reasoning process. Please see Figure~\ref{fig:case1} for an example, where MSSR correctly provides a detailed breakdown of its reasoning while GRPO fails. More case studies of reasoning outputs can be found in Appendix. 

% \textcolor{red}{[todo]fill the total response length for each example in Figure 5.  [e.g., total words: xxx]}

\begin{table}[h!]
\centering
\small
% \vspace{-8pt}
\caption{\textbf{Comparison of reasoning granularity.} 
We measure the average number of key reasoning steps and average response length on the MMK12 benchmark. MSSR produces more fine-grained reasoning (3.3 key steps on average) compared to GRPO (1.9) and the base model (3.1), indicating more robust, well-structured reasoning. The trend is consistent with average response length, where MSSR outputs are longer and more detailed than those from GRPO.}

\begin{tabular}{lccc}
\toprule
Reasoning Granularity              & BASE-7B & GRPO & MSSR \\
\midrule
Average number of key steps        & 3.1  & 1.9  & 3.3 \\
Average response length  & 488.2 & 439.7  & 511.6 \\
\bottomrule
\end{tabular}
\label{tab:key_steps}
\end{table}

%% file: sec/conclusions.tex
\vspace{-10pt}
\section{Conclusions}
This paper tackles the fundamental trade-off between training compute efficiency and training stability in RLVR for multimodal reasoning. We show that group-based methods like GRPO are compute-inefficient due to multi-rollout sampling, while naive single-rollout approaches suffer from instability and entropy collapse. To address this, we introduce MSSR, a group-free framework that achieves both efficiency and stability through entropy-based advantage shaping, which adaptively regularizes advantage magnitudes to prevent entropy collapse and stabilize multimodal training. Extensive experiments demonstrate that MSSR matches the validation performance of strong GRPO baselines with half of the total training steps  and consistently outperforms them under similar budgets across five multimodal reasoning benchmarks. Ablation studies further reveal that alternative explored stabilization strategies are insufficient, showing that entropy-based advantage shaping is critical for stabilizing single-rollout multimodal reasoning. % (highlighting its superior compute efficiency)

% This paper addresses the critical trade-off between sample efficiency and training stability in RLVR for multimodal reasoning. We first demonstrate that while group-based methods like GRPO are sample inefficient due to their multi-rollout sampling, naive single-rollout approaches suffer from severe instability and entropy collapse in multimodal settings. To overcome this, we proposed MSSR, a group-free framework that achieves both sample efficiency and stability. The core of MSSR is an entropy-based advantage-shaping mechanism that adaptively regularizes advantage magnitudes, preventing entropy collapse and promoting stable exploration. 

% Our extensive experimental results show that MSSR is highly effective. It is more sample-efficient, matching the performance of the strong GRPO baseline in less than half the training steps. When trained for a comparable budget, MSSR consistently surpasses group-based methods in both in-distribution accuracy and generalization performance across five diverse multimodal reasoning benchmarks. Furthermore, our ablation studies confirmed that other non-trivial stabilization strategies, such as cross-modal regularization or a simple entropy loss, were insufficient. This highlights our key finding: entropy-based advantage shaping is not merely beneficial but essential for stabilizing the multimodal single-rollout setting.

% Overall, our findings demonstrate that MSSR alleviates the sample efficiency-stability dilemma, enabling stable, sample-efficient, and effective RLVR for advancing complex multimodal reasoning tasks.

%% file: sec/suppl.tex
\clearpage
\setcounter{page}{1}
\maketitlesupplementary

\subsection{Training Cost}

We measure the training cost for both 3B and 7B models in terms of minutes per training step. As shown in Table~\ref{tab:time}, MSSR maintains a per-step cost similar to GRPO, introducing only minimal overhead due to the initial baseline estimation, while requiring far fewer total steps to achieve the final performance level of GRPO, as shown in Figure \ref{fig:page1}.

\begin{table}[ht]
    \centering
    \caption{Training cost across methods on Qwen2.5-VL 3B and 7B models, measured as the average training time per step (mins/step). }
    \begin{tabular}{lc}
        \toprule
        \textbf{Model} & \textbf{\makecell{Train Cost \\ (mins/step)}} \\
        \midrule
        Qwen2.5-VL-3B \citep{Qwen-VL}  & --\\
        \quad + GRPO \cite{shao2024deepseekmath} & 3.9\\ 
        \quad + RLOO \cite{ahmadian2024back} & 3.9 \\ 
        \quad + REINFORCE++ \cite{hu2025reinforce++} & 4.8 \\ 
        \quad + MSSR & 3.9\\
        \midrule
        Qwen2.5-VL-7B \citep{Qwen-VL}& --\\
        \quad + GRPO \cite{shao2024deepseekmath} & 6.1 \\ 
        \quad + RLOO \cite{ahmadian2024back} & 6.2 \\ 
        \quad + REINFORCE++ \cite{hu2025reinforce++}  & 8.3\\
        \quad + MSSR & 6.9 \\
        \bottomrule
    \end{tabular}
    \label{tab:time}
\end{table}

\subsection{Training Accuracy for Sensitivity Analysis}
\begin{figure*}[t]
    \centering
    \begin{subfigure}[t]{0.4\linewidth}
        \centering
        \includegraphics[width=\textwidth]{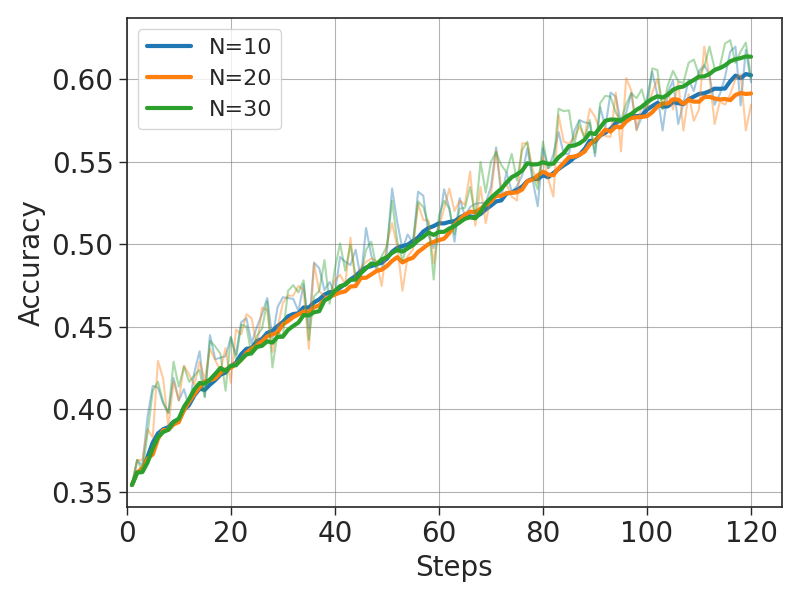}
        \caption{Training accuracy for varying $N$}
        \label{fig:n_train}
    \end{subfigure}
    % \begin{subfigure}[t]{0.49\linewidth}
    %     \centering
    %     \includegraphics[width=\textwidth]{figs/n_val_pass@1_7b.png}
    %     \caption{Validation accuracy}
    %     \label{fig:n_val}
    % \end{subfigure}
\hspace{3pt}
    \begin{subfigure}[t]{0.4\linewidth}
        \centering
        \includegraphics[width=\textwidth]{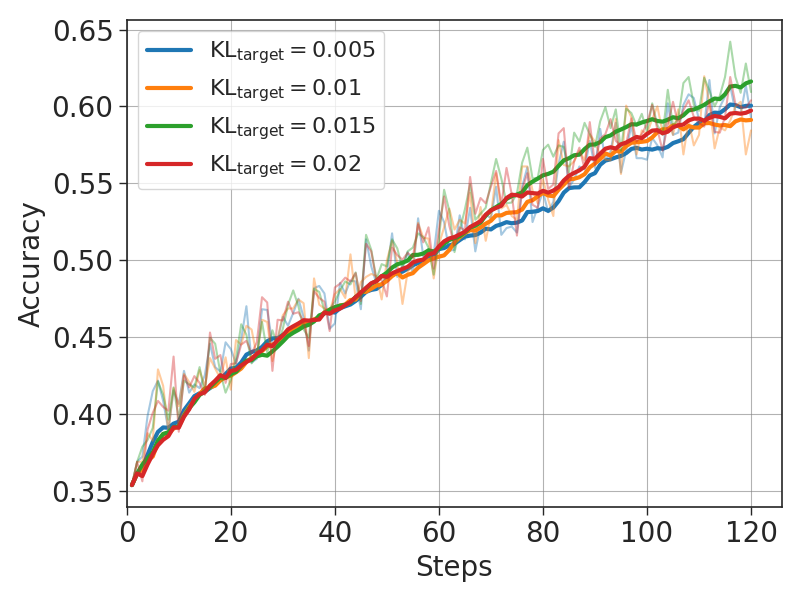}
        \caption{Training accuracy for varying $\mathrm{KL}_{\text{target}}$}
        \label{fig:kl_train}
    \end{subfigure}
    % \begin{subfigure}[t]{0.49\linewidth}
    %     \centering
    %     \includegraphics[width=\textwidth]{figs/kl_val_pass@1_7b.png}
    %     \caption{Validation accuracy}
    %     \label{fig:kl_train}
    % \end{subfigure}
    \caption{\textbf{Sensitivity analysis for effects of sliding window size $N$ and target KL value $\mathrm{KL}_{\text{target}}$ on adjusting the discount factor $\eta$ and their impact on training accuracy.}
(a) Training accuracy under different sliding window sizes $N$.
(b) Training accuracy under different target KL values $\mathrm{KL}_{\text{target}}$. Across all settings, the training accuracy curves remain largely similar.}
    \label{fig:sen_train}
\end{figure*}

In Section~\ref{sec:sensitivity}, we study how the sliding window size $N$ and the target KL divergence $\text{KL}_\text{target}$ affect validation accuracy. We find that $N=20$ and $\text{KL}_\text{target}=0.01$ yield the strongest performance. For completeness, we also report the corresponding training accuracy curves in Figure~\ref{fig:sen_train}. The training trends across different choices of $N$ and $\text{KL}_\text{target}$ remain broadly similar.

\subsection{Performance Comparison}
We compare MVSR and MSSR on Qwen2.5-VL 3B and 7B models for their generalization performance on diverse multimodal reasoning benchmarks. We present the results in Table \ref{tab:main_2}. Due to the training instability, MVSR provides limited gains over the base model for the 7B setting, and for the 3B model, its fine-tuned performance is often lower than the base model across most benchmarks. In contrast, MSSR consistently improves generalization performance across all benchmarks.

\begin{table*}[t]
    \centering
    \caption{\textbf{Model generalization performance on diverse multimodal reasoning benchmarks.} We compare MVSR and MSSR on Qwen2.5-VL 3B and 7B models. As discussed earlier, MVSR's training instability leads to limited gains for the 7B model and, for the 3B model, often results in performance below the base model across most benchmarks. In contrast, MSSR consistently improves generalization performance across all benchmarks.}

    \resizebox{\textwidth}{!}{
    \begin{tabular}{lcccccc}
        \toprule
        \textbf{Model} & \textbf{MathVerse} & \textbf{MathVista}  & \textbf{MMK12} & \textbf{\makecell{R1-Onevision \\ Bench}} & \textbf{HallusionBench}  & \textbf{Avg.}  \\
        \midrule
        Qwen2.5-VL-3B \citep{Qwen-VL} & 33.3 & 59.5   & 42.5 & 27.6 & 59.9  & 44.6 \\
        \quad + MVSR & 31.3 & 52.9  & 41.3 & 15.2 & 64.8  & 41.1  \\ 
        \quad + MSSR  & \textbf{39.6} & \textbf{63.0}  & \textbf{49.2} & \textbf{29.0} & \textbf{66.6}  & \textbf{49.5} \\
        \midrule
        Qwen2.5-VL-7B \citep{Qwen-VL} & 45.8 & 67.2   & 48.1 & 34.6 & 68.4 & 52.8 \\
        \quad + MVSR & 43.9 & 69.0  & 51.7 & 35.5 & 68.7  & 53.8 \\
        \quad + MSSR & \textbf{49.8} & \textbf{71.1}   & \textbf{62.5} & \textbf{39.2} & \textbf{70.6}  & \textbf{58.6}  \\
        \bottomrule
    \end{tabular}
    }
    \label{tab:main_2}
\end{table*}

% \subsection{Additional Experiments}
% We conduct additional experiments by training more steps to see if MSSR is still stable. We training for 150 steps for the approach of MVSR and MSSR on the Qwen2.5-VL-7B mdoel. As shown in Figure 

% \begin{figure*}[t]
%     \centering
%     \begin{subfigure}[t]{0.4\linewidth}
%         \centering
%         \includegraphics[width=\textwidth]{figs/n_reward_accuracy_7b.png}
%         \caption{Training accuracy for 150 steps}
%         \label{fig:150_train}
%     \end{subfigure}
% \hspace{4pt}
%     \begin{subfigure}[t]{0.4\linewidth}
%         \centering
%         \includegraphics[width=\textwidth]{figs/kl_reward_accuracy_7b.png}
%         \caption{Validation accuracy for 150 steps }
%         \label{fig:150_val}
%     \end{subfigure}

%     \caption{\textbf{Sensitivity analysis for effects of sliding window size $N$ and target KL value $\mathrm{KL}_{\text{target}}$ on adjusting the discount factor $\eta$ in baseline estimation and their impact on training accuracy.}
% (a) Training accuracy under different sliding window sizes $N$.
% (b) Training accuracy under different target KL values $\mathrm{KL}_{\text{target}}$. Across all settings, the training accuracy curves remain largely similar.}
%     \label{fig:150_steps}
% \end{figure*}

\subsection{Reasoning Case Studies}
In Section~\ref{sec: exp} Experiments, we present a representative case study in Figure~\ref{fig:case1} comparing the reasoning outputs of GRPO and MSSR. MSSR arrives at the correct solution, whereas GRPO fails. We mark the erroneous reasoning steps of GRPO in red and the key steps enabling MSSR’s correct prediction in green.

\begin{figure*}[t]
    \centering
    \includegraphics[width=\textwidth]{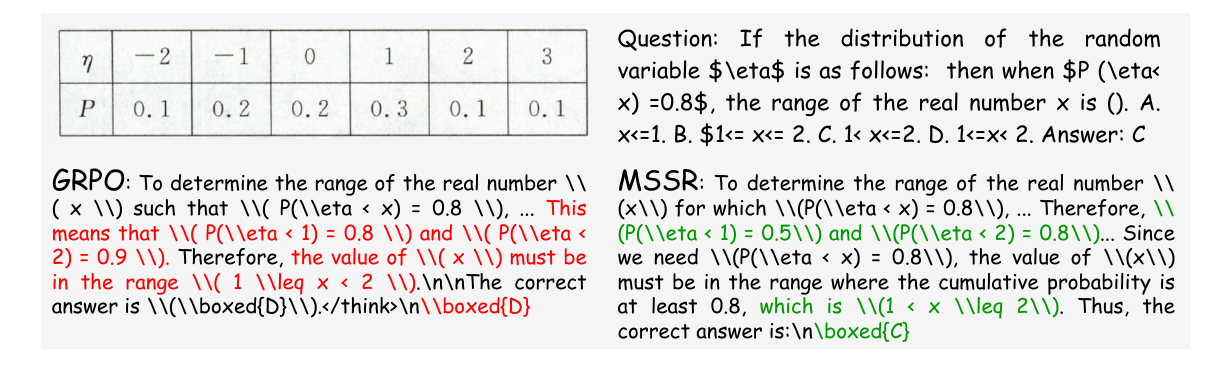}
    \caption{\textbf{Example of reasoning outputs.} Comparing the Qwen2.5-VL-7B model fine-tuned with GRPO and MSSR. While GRPO produce incorrect answers, MSSR successfully solves the problem, demonstrating its superior reasoning capability. We highlight the critical reasoning steps that lead to GRPO's incorrect answer in \textcolor{red}{red}, and the key steps enabling MSSR’s correct prediction in \textcolor{ForestGreen}{green}.}
    \label{fig:case2}
    % \vspace{-10pt}
\end{figure*}

\begin{figure*}[t]
    \centering
    \includegraphics[width=\textwidth]{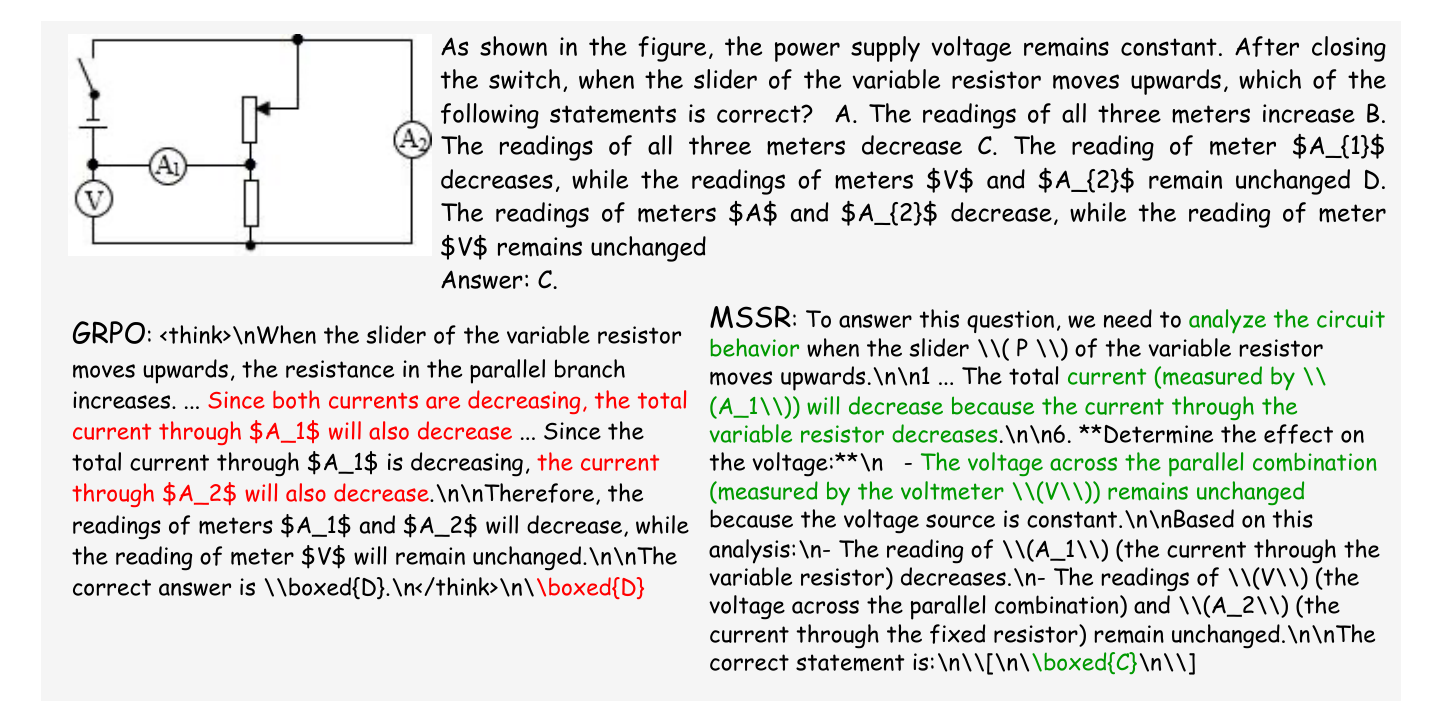}
    \caption{\textbf{Example of reasoning outputs.} Comparing the Qwen2.5-VL-7B model fine-tuned with GRPO and MSSR. While GRPO produce incorrect answers, MSSR successfully solves the problem, demonstrating its superior reasoning capability. We highlight the critical reasoning steps that lead to GRPO's incorrect answer in \textcolor{red}{red}, and the key steps enabling MSSR’s correct prediction in \textcolor{ForestGreen}{green}.}
    \label{fig:case3}
\end{figure*}

\begin{figure*}[t]
    \centering
    \includegraphics[width=0.7\textwidth]{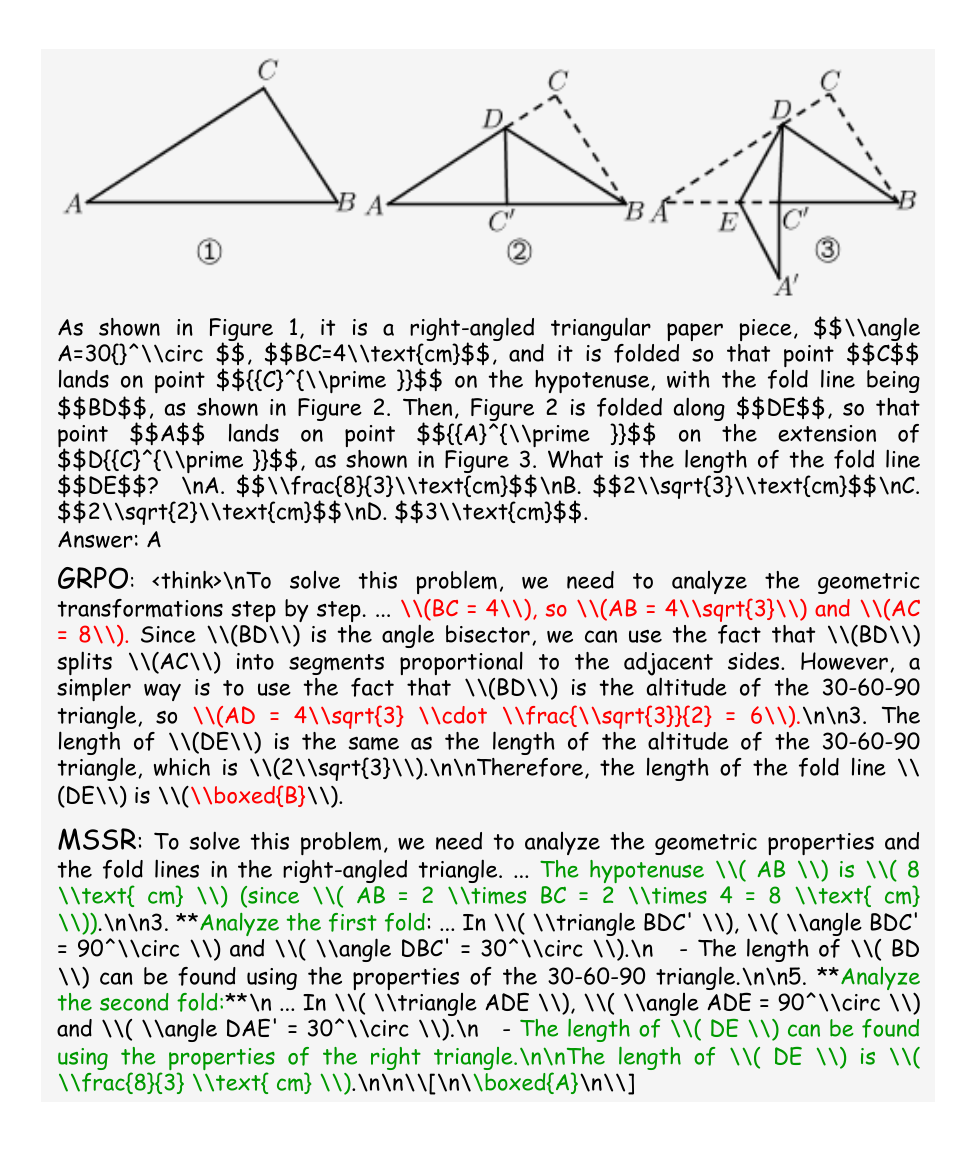}
    \caption{\textbf{Example of reasoning outputs.} Comparing the Qwen2.5-VL-7B model fine-tuned with GRPO and MSSR. While GRPO produce incorrect answers, MSSR successfully solves the problem, demonstrating its superior reasoning capability. We highlight the critical reasoning steps that lead to GRPO's incorrect answer in \textcolor{red}{red}, and the key steps enabling MSSR’s correct prediction in \textcolor{ForestGreen}{green}.}
    \label{fig:case4}
\end{figure*}

For illustration, we provide additional qualitative examples in Figures~\ref{fig:case2}, \ref{fig:case3}, and \ref{fig:case4}. Across these cases, GRPO gives incorrect answers, while MSSR successfully solves the problems, further showcasing its stronger and more reliable reasoning capability.

\subsection{Prompt Templates} \label{app: prompt}
We list below the prompt used to instruct the model to produce the structured outputs.
\begin{tcolorbox}[colback=gray!10, colframe=gray!70, title=System Prompt]
You FIRST think about the reasoning process as an internal monologue and then provide the final answer. The reasoning process MUST be enclosed within \(\texttt{<think></think>}\) tags. The final answer MUST be put in \texttt{\textbackslash boxed\{\}}.
\end{tcolorbox}

% \begin{figure*}[ht]
%     \centering
%     \includegraphics[width=\textwidth]{figs/MLLM_Exploration-Page-7.drawio.pdf}
%     \caption{\textbf{Example of reasoning outputs.} Comparing the Qwen2.5-VL-7B model fine-tuned with GRPO, MVSR, and MSSR. While GRPO and MVSR produce incorrect answers, MSSR successfully solves the problem, demonstrating its superior reasoning capability. We highlight the critical reasoning steps that lead to other baselines' incorrect answer in \textcolor{red}{red}, and the key steps enabling MSSR’s correct prediction in \textcolor{ForestGreen}{green}.}
%     \label{fig:case2}
%     % \vspace{-10pt}
% \end{figure*}

% \begin{figure*}[ht]
%     \centering
%     \includegraphics[width=\textwidth]{figs/MLLM_Exploration-Page-7.drawio.pdf}
%     \caption{\textbf{Example of reasoning outputs.} Comparing the Qwen2.5-VL-7B model fine-tuned with GRPO, MVSR, and MSSR. While GRPO and MVSR produce incorrect answers, MSSR successfully solves the problem, demonstrating its superior reasoning capability. We highlight the critical reasoning steps that lead to other baselines' incorrect answer in \textcolor{red}{red}, and the key steps enabling MSSR’s correct prediction in \textcolor{ForestGreen}{green}.}
%     \label{fig:case2}
%     % \vspace{-10pt}
% \end{figure*}